\documentclass[sigconf]{acmart}

\usepackage{bbding}
\usepackage{fancyhdr}

\AtBeginDocument{%
	}

\copyrightyear{2024}
\acmYear{2024}
\setcopyright{rightsretained}
\acmConference[MM '24]{Proceedings of the 32nd ACM International Conference on Multimedia}{October 28-November 1, 2024}{Melbourne, VIC, Australia}
\acmBooktitle{Proceedings of the 32nd ACM International Conference on Multimedia (MM '24), October 28-November 1, 2024, Melbourne, VIC, Australia}
\acmDOI{10.1145/3664647.3681226}
\acmISBN{979-8-4007-0686-8/24/10}

\makeatletter
\gdef\@copyrightpermission{
	\begin{minipage}{0.3\columnwidth}
		\href{https://creativecommons.org/licenses/by-nc/4.0/}{\includegraphics[width=0.90\textwidth]{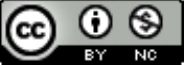}}
	\end{minipage}\hfill
	\begin{minipage}{0.7\columnwidth}
		\href{https://creativecommons.org/licenses/by-nc/4.0/}{This work is licensed under a Creative Commons Attribution-NonCommercial International 4.0 License.}
	\end{minipage}
	\vspace{5pt}
}
\makeatother

\begin{document}
	
	\title{Cefdet: Cognitive Effectiveness Network Based on Fuzzy Inference for Action Detection}
	
	
	
	\author{Zhe Luo}
	\orcid{0009-0005-8614-2847}
	\affiliation{%
		\institution{Hunan Normal University}
		\streetaddress{No.36, Lushan Road}
		\city{Changsha}
		\state{}
		\country{China}
		\postcode{410081}
	}
	\email{luozhe@hunnu.edu.cn}
	
	\author{Weina Fu}
	\orcid{0000-0002-4302-6505}
	\affiliation{%
		\institution{Hunan Normal University}
		\streetaddress{No.36, Lushan Road}
		\city{Changsha}
		\country{China}}
	\email{fuwn@hunnu.edu.cn}

	\author{Shuai Liu}
	\orcid{0000-0001-9909-0664}
	\authornote{Corresponding author}
	\affiliation{%
		\institution{Hunan Normal University}
		\city{Changsha}
		\country{China}
	}
	\email{liushuai@hunnu.edu.cn}

	\author{Saeed Anwar}
	\orcid{0000-0002-0692-8411}
	\affiliation{%
		\institution{The Australian National University}
		\city{Canberra}
		\country{Australia}}
	\email{saeed.anwar@anu.edu.au}
	
	\author{Muhammad Saqib}
	\orcid{0000-0003-4374-0888}
	\affiliation{%
		\institution{CSIRO NCMI}
		\city{Marsfield Sydney}
		\country{Australia}
	}
	\email{muhammad.saqib@csiro.au}
	
	\author{Sambit Bakshi}
	\orcid{0000-0002-6107-114X}
	\affiliation{%
		\institution{National Institute of Technology}
		\city{Rourkela}
		\country{India}
	}
	\email{bakshisambit@ieee.org}
	
	\author{Khan Muhammad}
	\orcid{0000-0002-5302-1150}
	\authornote{Corresponding author}
	\affiliation{%
		\institution{Sungkyunkwan University}
		\city{Seoul}
		\country{Republic of Korea}
	}
	\email{khanmuhammad@g.skku.edu}


	\renewcommand{\shortauthors}{Zhe Luo et al.}
	
	\begin{abstract}
		Action detection and understanding provide the foundation for the generation and interaction of multimedia content. However, existing methods mainly focus on constructing complex relational inference networks, overlooking the judgment of detection effectiveness. Moreover, these methods frequently generate detection results with cognitive abnormalities. To solve the above problems, this study proposes a cognitive effectiveness network based on fuzzy inference (Cefdet), which introduces the concept of “cognition--based detection” to simulate human cognition. First, a fuzzy--driven cognitive effectiveness evaluation module (FCM) is established to introduce fuzzy inference into action detection. FCM is combined with human action features to simulate the cognition--based detection process, which clearly locates the position of frames with cognitive abnormalities. Then, a fuzzy cognitive update strategy (FCS) is proposed based on the FCM, which utilizes fuzzy logic to re--detect the cognition--based detection results and effectively update the results with cognitive abnormalities. Experimental results demonstrate that Cefdet exhibits superior performance against several mainstream algorithms on the public datasets, validating its effectiveness and superiority. Code is available at https://github.com/12sakura/Cefdet.
	\end{abstract}

	
	
	
	\maketitle
	
	\section{Introduction}
	Multimedia is an essential branch of modern information technology, which comprehensively processes various media forms of information such as text, audio, and video. With the improvement of computer processing capabilities, the creation and sharing of multimedia content have been further promoted. Multimedia technology has driven the innovation of content presentation and provides a wealth of application scenarios for artificial intelligence research \cite{fan2021motion, liu2023weakly}.
	
	Action detection is widely applied in multimedia technology. It focuses on identifying specific human actions or behavior patterns from videos. In recent years, the accuracy of action detection has significantly improved with the advent of deep learning, particularly convolutional neural networks (CNNs) and recurrent neural networks (RNNs). Action detection is gradually playing a crucial role in various applications, including smart security, health monitoring, and human--computer interactions \cite{chen2021learning, liu2024two}.
	
	\begin{figure}[htb]%
		\centering
		\includegraphics[width=1\linewidth]{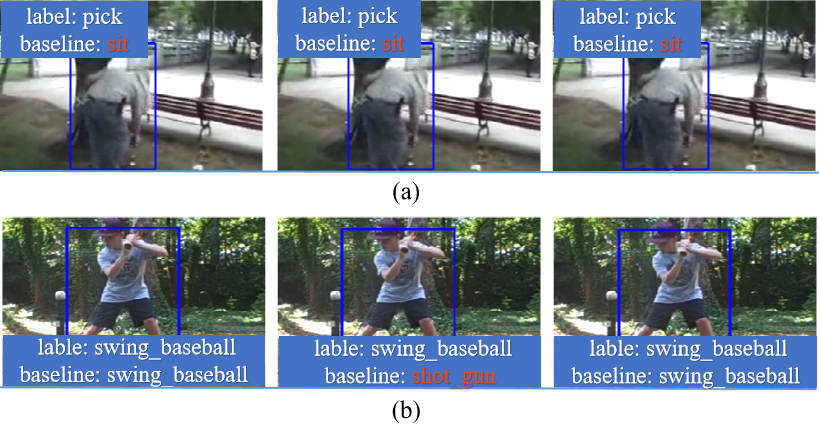}
		\caption{Detection results with cognitive abnormalities in three consecutive frames. (a) is the false detection result of existing methods in highly similar actions, and (b) denotes the detection results of existing methods that do not conform to human action norms.}
		\label{fig:fig1}
	\end{figure}
	
	However, existing methods face challenges in determining the effectiveness of detection results. Additionally, existing methods constantly generate detection results with cognitive abnormalities due to ineffective judgment. As shown in Figure \ref{fig:fig1} (a), actions such as “pick” and “sit” exhibit high similarity. Existing methods wrongly detect “pick” as “sit” while affecting the action detection of subsequent frames. In addition, as depicted in Figure \ref{fig:fig1} (b), action changes from “swing\_baseball” to “shot\_gun” do not conform to human action norms. Existing methods detect two unrelated actions as adjacent and recognize one or several consecutive frames as a complete action, which is inconsistent with human cognition. These problems hinder the further application of action detection methods.
	
	Therefore, this study proposes a cognitive effectiveness network based on fuzzy inference (Cefdet). It introduces the concept of “cognition--based detection” and combines features in human action with fuzzy inference to simulate the cognition--based detection process. The effectiveness of the detection results is accurately determined using cognition--based detection. Moreover, the detection results are divided based on their effectiveness, and fuzzy logic is employed to dynamically update the detection results, which repairs the detection results with cognitive abnormalities. 
	
	The contributions of this study are summarized as follows:
	
	\begin{itemize}
		\item This study proposes the FCM for detecting the effectiveness of frames. It takes the features of human actions, including the confidence of each frame, the correlation between adjacent frames, and the position score of each frame, as inputs to the fuzzy inference system. This simulates a cognition--based detection process to obtain effectiveness, thus accurately locating the position of action frames with cognitive abnormalities.
		\item In this study, the FCS based on the above FCM is employed to update the results dynamically. It divides cognition--based detection results into high and low--levels. Subsequently, the local features of the frames with low--level cognition are weighted through frames with high--level cognition for re--detection. The frames before and after re--detection are dynamically updated to obtain optimal results, effectively repairing the detection results with cognitive abnormalities.
		\item Experimental results show that Cefdet  performs better on public datasets than existing methods. Furthermore, Cefdet promotes the application of fuzzy inferences in computer vision perception.
	\end{itemize}
	

	\section{Related work}
	
	\subsection{Video understanding}
	
	Video understanding includes identifying the activities that occur during the editing process. Typically, the time span for editing is a few seconds, and there is only one annotation. Most early video comprehension methods \cite{wang2016temporal, zhou2018temporal} utilize 2D image CNNs and introduced long--short--term memory (LSTM) \cite{donahue2015long} to learn video time structures. These methods are difficult to capture the rich spatiotemporal information in videos. Some studies \cite{qiu2017learning, duan2022revisiting} have attempted to use 3D CNNs to model these complex spatiotemporal information and have achieved excellent results. Two--stream networks \cite{simonyan2014two} constitute another widely employed video understanding method. Due to its ability to handle only a small portion of the input frames, it achieves a superior balance between accuracy and complexity. Recently, Cheng et al. \cite{cheng2023sample} proposes a new method to recover the intermediate features between two sparse samples and adjacent video frames. Singh et al. \cite{singh2023eval} constructs advanced positional correlation models for any specific scenario by learning the general representation of motion to improve its interpretability. Bao et al. \cite{bao2022hierarchical} are dedicated to the study of static and dynamic backgrounds, effectively integrating "reconstruction" and "prediction" tasks through the binding mechanism between foreground and background.
	
	\subsection{Action detection}
	
	Action detection has received increasing attention from researchers as an essential technology for video understanding \cite{feichtenhofer2019slowfast}. Due to the potential of multiple actions within each frame, it is necessary to detect the actions of individual entities within the current frame rather than categorizing the entire video into a single class. Inspired by deep CNNs for object detection \cite{peng2016multi}, methods based on action detection typically apply 2D position anchors or offline object detectors to key--frames to locate human subjects \cite{wu2020context, zhao2022tuber}. They focus on improving action detection and incorporating temporal patterns by leveraging the optical flow for additional flow fusion. In addition, some methods \cite{gu2018ava, sun2018actor} apply 3D convolutional networks to capture temporal information for identifying actions and achieve excellent results.
	
	Recent research on spatiotemporal action detection emphasizes modeling the interaction between classified individuals and their environment. Recent methods \cite{tang2020asynchronous, pan2021actor} propose explicitly modeling the relationships between actors and objects. A dual--mode interaction structure \cite{faure2023holistic} is constructed based on human posture, hands, and objects, effectively improved the accuracy of action detection. In addition, Guan et al. \cite{guan2023egocentric} designs a two-stage training scheme to model the correlation between video segments. Liu et al. \cite{liu2024knowledge} constructs a knowledge-based hierarchical causal inference network, which improves the performance of action recognition and the interpretability of the model by comparing the results of factual reasoning and counterfactual reasoning.
	
	\subsection{Fuzzy inference}
	
	Fuzzy inference is widely employed in computer vision applications. It is a core component of fuzzy logic and involves the application of fuzzy sets and rules to input data to obtain fuzzy output results. Fuzzy inference performs better in handling uncertain information and can more reasonably describe the actual needs. Several studies based on fuzzy inference are increasingly growing. For example, an improved fuzzy clustering--based classifier \cite{kim2017design} employs $L_{2}$ regularization to mitigate overfitting. This classifier demonstrates remarkable performance in various classification tasks. Liu et al. \cite{liu2020fuzzy} introduces a fuzzy detection strategy to prejudge detection results and improves detection robustness in complex environments. Multi--layer interval Type--2 fuzzy limit machine learning (ML--IT2--FELM) \cite{rubio2020multilayer} recognizes walking activities and uses wearable sensors to determine gait. Fuzzy learning is utilized to obtain appropriate results for the activities. Cao et al. \cite{cao2020multiobjective} integrates interval Type--2 fuzzy sets into a fuzzy rough neural network to forecast stock time series. In addition, other studies \cite{li2017optimal, rong2018finite} introduce a common quadratic Lyapunov function to analyze the stability and design controllers for fuzzy closed--loop systems.

	\section{Proposed method}
	
	\subsection{Overall framework}
	
	\begin{figure*}[htb]%
		\centering
		\includegraphics[width=1\linewidth]{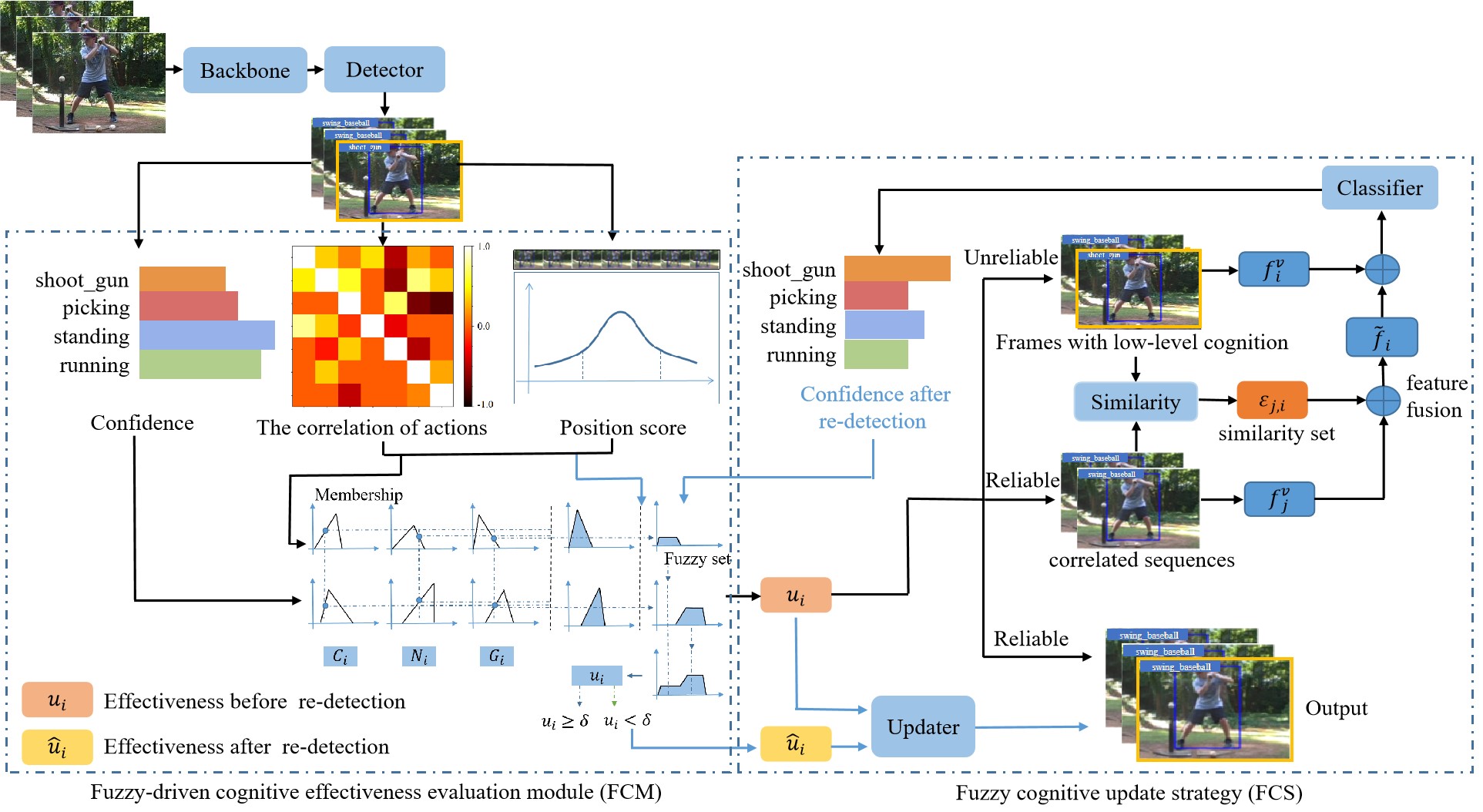}
		\caption{The overall framework of Cefdet for action detection. The left module is the Fuzzy--driven cognitive effectiveness evaluation module, abbreviated as FCM, and on the right is the Fuzzy cognitive update strategy, termed FCS.}
		\label{fig:fig2}
	\end{figure*}
	
	Due to the uncertainty of daily human activities, existing methods cannot effectively determine the effectiveness of their detection results. Moreover, the lack of effective judgment frequently leads to detection results with cognitive abnormalities in existing methods. To solve these problems, this study introduces the concept of “cognition--based detection” into action detection. First, fuzzy inference is employed to simulate the cognition--based detection process by combining human action features to accurately locate the positions of frames with cognitive abnormalities. Then, the local features of the frames with cognitive abnormalities are dynamically weighted based on fuzzy logic for re--detection. The effectiveness before and after re--detection is combined to repair detection results inconsistent with cognition.
	
	The overall framework of Cefdet is illustrated in Figure \ref{fig:fig2}. Initially, the FCM is designed after the video frames pass through a network. It evaluates the effectiveness of each frame by considering the confidence, the correlation between adjacent frames, and the position score of each frame, dividing the frames into high and low--levels. In addition, the FCS is proposed to update the frames with low--level cognition. Specifically, it constructs correlated sequences using frames with high--level cognition. Then, the features are weighted based on the similarity between the correlated sequences and frames with low--level cognition for re--detection. Finally, an updater based on the effectiveness before and after the re--detection is designated for updating the results.

	\subsection{Fuzzy--driven cognitive effectiveness evaluation module (FCM)}
	Due to the lack of effective judgment of detection results, existing action detection methods frequently obtain detection results with cognitive abnormalities in complex scenes. The use of confidence scores to evaluate the effectiveness of an action frame is biased, and more factors should be considered. Therefore, this study proposes the FCM that cooperatively judges each frame's effectiveness by combining each frame's confidence, the correlation between adjacent frames, and the position score of each frame with a fuzzy inference system. Fuzzy logic is employed to simulate the cognition--based detection process to accurately determine the effectiveness of each frame and locate the positions of frames with cognitive abnormalities. FCM is a fuzzy inference engine with four components: feature quantization, fuzzification, fuzzy inference, and defuzzification. Brief descriptions of each component are provided below.
	
	\noindent
	\textbf{Feature quantification.} The objective of fuzzy inference is to assess the effectiveness of the frames after detection. Therefore, it is essential to construct feature vectors of the frames as inputs for fuzzy inference. For the input video sequence $I_{i}$, the feature vector consists of diverse feature values, including the confidence of each frame $C_{i}$, the correlation between adjacent frames $N_{i}$, and the position score of each frame $G_{i}$.
	
	1. The confidence of each frame. As an important evaluation criterion, the confidence of the action detection network is expressed in the following formula.
	
	\begin{equation}
		C_{i} = \varphi\left( I_{i} \right),
	\end{equation}
	
	\noindent
	where $\varphi(*)$ refers to the action detection network in \cite{faure2023holistic}, which is tasked to predict the true value from $I_{i}$.
	
	\begin{figure}[htb]%
		\centering
		\includegraphics[width=0.8\linewidth]{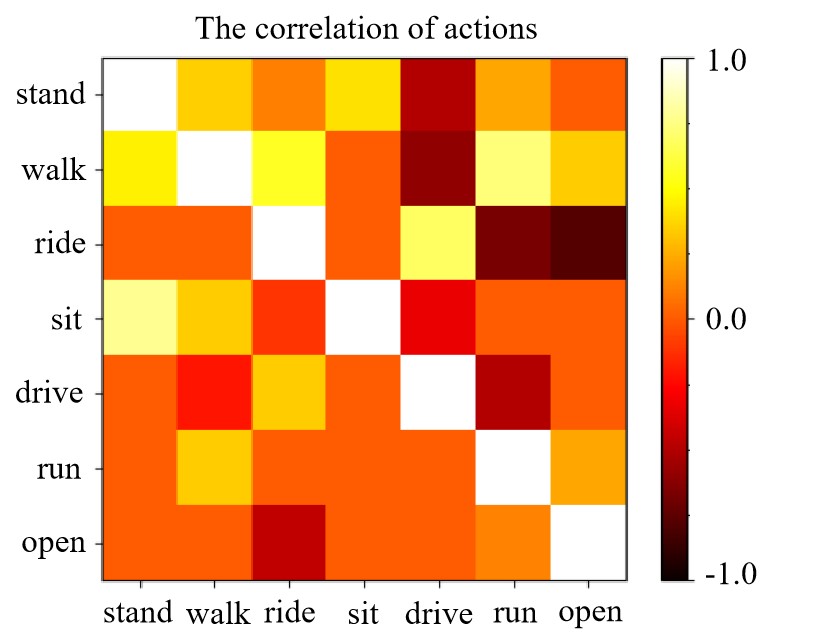}
		\caption{The correlation between different actions. The correlations between all actions are normalized to be between [-1, 1], with darker colors indicating lower correlations.}
		\label{fig:fig3}
	\end{figure}
	
	2. The correlation between adjacent frames. In video sequences, action sequences with short intervals have rich correlations, and the evolution from one action to another is a gradual process. It is possible to determine whether the subsequent action is reasonable based on a previous action. Therefore, the positions of action frames that do not conform to human action norms are accurately located using this correlation. This study employs normalized pointwise mutual information (NPMI) to measure the correlation between actions. NPMI is defined as follows:
	
	\begin{equation}
		N_{i} = \left( ln\frac{P\left( {K_{I_{i - 1}},K_{I_{i}}} \right)}{P\left( K_{I_{i - 1}} \right)p\left( K_{I_{i}} \right)} \right) \slash \left( - {\ln{P\left( {K_{I_{i - 1}},K_{I_{i}}} \right)}} \right),
	\end{equation}
	
	\noindent
	where $K_{I_{i}}$ refers to the action category of the i--th frame; $P\left( K_{I_{i}} \right)$ is the probability of $K_{I_{i}}$; and $P\left( {K_{I_{i - 1}},K_{I_{i}}} \right)$ denotes the probability of a joint distribution between actions. The value of $N_{i}$ is in the range [-1, 1], where $N_{i} = 1$ represents a high degree of correlation between the two actions, $N_{i} = 0$ indicates independence between the two actions, and $N_{i} = -1$ denotes that the two actions have never appeared simultaneously. Figure \ref{fig:fig3} shows the correlation between the different actions calculated by the NPMI.
	
\begin{figure}[htb]%
		\centering
		\includegraphics[width=0.8\linewidth]{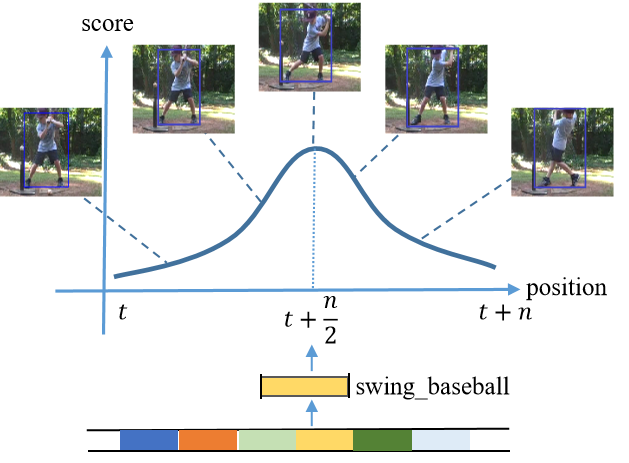}
		\caption{The position score of frames in action sequences. In a complete action sequence, the closer to the center the frame is, the more reliable it is.}
		\label{fig:fig4}
	\end{figure}

	3. The position score of each frame. Action detection is the process of observation and refinement with a clear time boundary between the beginning and end of an action. The position of the current action frame is noted. If the frame is at the center of the current action, it is considered highly reliable. By contrast, if the frame is at the boundary position of the current action, the reliability of the frame requires further detection. In this study, the reliability of a frame is determined by comparing its corresponding position within a complete action. For action category $K_{I_{t}}$ at frame $t$, if $K_{I_{t - 1}} \neq K_{I_{t}},K_{I_{t}} = K_{I_{t + 1}} = \cdots = K_{I_{t + n}}$ and $K_{I_{t}} \neq K_{I_{t + n + 1}}$, then $\left\lbrack I_{t},\cdots,I_{t + n} \right\rbrack$ can be considered as a complete action. The reliability of the frame is defined as follows:
	
	\begin{equation}
		G_{i} = \frac{1}{\sigma\sqrt{2\pi}}e^{- \frac{1}{2}z^{2}},
	\end{equation}
	
	\begin{equation}
		z = \frac{i - \mu}{\sigma},~i \in \lbrack t,t + n\rbrack,
	\end{equation}
	
	\noindent
	where $G_{i}$ is the position score of the i--th frame; $\mu$ and $\sigma$ are the mean and standard deviation of the Gaussian distribution, respectively; $\mu = t + \frac{n}{2}$, and $\sigma = \sqrt{\frac{\sum_{i = t}^{t + n}(i - \mu)^{2}}{n}}$. Figure \ref{fig:fig4} shows the position scores of frames in the complete action sequence.
	
	\begin{sloppypar}
		\noindent
		\textbf{Fuzzification.} The five--level fuzzification rule $k = \left\{ NB,~NS,ZO,PS,PB \right\}$ is employed to determine the degree of membership of the feature vector to a fuzzy set: negative big ($NB$), negative small ($NS$), zero ($ZO$), positive small ($PS$), or positive big ($PB$). $h_{i} = \left\{ C_{i},N_{i},G_{i},U_{i} \right\}$ is the training set of fuzzy inference, where $U_{i} = \mu_{1}C_{i} + \mu_{2}N_{i} + \left( 1 - \mu_{1} - \mu_{2} \right)G_{i}$, the universe is defined as $P = \left\{ P\left( h_{i} \right) \right\}$, $\mu_{k_{P}}\left( h_{i} \right)$ indicates the membership degree of $h_{i}$ to the set $k_{P}$, and $F_{h_{i}} = \left\{ \mu_{k_{P}}\left( h_{i} \right) \right\}$ reports that $h_{i}$ belongs to the fuzzy set of universe $P$, as illustrated below:
	\end{sloppypar}
	
	\begin{equation}
		F_{h_{i}} = \left\{ \frac{\mu_{k_{P(h_{i})}}\left( h_{i} \right)}{h_{i}} \right\}.
	\end{equation}
	
	Thus, the fuzzy control rule is obtained by extracting the front parts $F_{C_{i}}$, $F_{N_{i}}$, and $F_{G_{i}}$ and the rear part $F_{U_{i}}$.
	
	\noindent
	\textbf{Fuzzy inference.} The fuzzy inference system has three input variables, driving 125 fuzzy rules. It follows the fuzzy rule formulation in \cite{liu2020fuzzy} and combines human cognition of daily behavior to ensure the rationality and effectiveness of fuzzy rules. Examples of fuzzy rules are shown in Table \ref{tab:tab1}. The fuzzy set generated by each fuzzy rule $\omega_{i,j}$ is obtained by performing the combination of membership degrees on the front parts of each rule, where $j \in \lbrack 1,2,\cdots,w\rbrack$ is the number of active rules. Subsequently, the conclusion of fuzzy inference ${\hat{U}}_{i}$ is obtained through a disjunction operation. The formula is as follows:
	
	\begin{small} 
		\begin{equation}
			\omega_{i,j} = if\left( {C_{i}~is~F_{C_{i}}} \right)and\left( {N_{i}~is~F_{N_{i}}} \right)and\left( {G_{i}~is~F_{G_{i}}} \right)then\left( U_{i}^{j}~is~F_{U_{i}}^{j} \right),
		\end{equation}
	\end{small}
	
	\begin{equation}
		\widehat{U}_{i}=\bigvee_{j=1}^{w}\omega_{i,j},
	\end{equation}
	
	\noindent
	where $U_{i}^{j}$ is the j--th element in the output universe $P\left( U_{i} \right)$, and $F_{U_{i}}^{j} \in k_{P(U_{i})}$ denotes any fuzzy set on $P\left( U_{i} \right)$.
	
	\begin{table}[htbp]
		\centering
		\caption {Examples of partial fuzzy rules.}
		\label{tab:tab1}
		\begin{tabular}{lllll}
			\hline
			Rule  &  $C$  &  $N$  &  $G$  &  $U$   \\ \hline
			$R_{1}$  &  $NB$  &  $PB$  &  $NB$  &  $PB$   \\
			$R_{2}$  &  $NB$  &  $ZO$  &  $NS$  &  $PS$   \\
			$R_{3}$  &  $NS$  &  $NS$  &  $ZO$  &  $ZO$   \\
			$R_{4}$  &  $PS$  &  $NB$  &  $PB$  &  $NS$   \\
			$R_{5}$  &  $NS$  &  $ZO$  &  $ZO$  &  $PS$   \\
			$R_{6}$  &  $NS$  &  $NS$  &  $NB$  &  $ZO$   \\
			$R_{7}$  &  $PB$  &  $NB$  &  $PB$  &  $NS$   \\
			$R_{8}$  &  $NS$  &  $NS$  &  $ZO$  &  $NB$   \\ \hline
		\end{tabular}
	\end{table}
	
	\noindent
	\textbf{Defuzzification.} During defuzzification, the centroid method is selected to obtain the output. The abscissa value corresponding to the membership function curve of the fuzzy set and the center of the area surrounded by it is the effectiveness of frame $u_{i}$, which can be defined as follows:
	
	\begin{equation}
		u_{i} = \frac{{\sum_{j = 1}^{w}U_{i}^{j}}\mu_{k_{P{(U_{i})}}}\left( U_{i}^{j} \right)}{\sum_{j = 1}^{w}{\mu_{k_{P(U_{i})}}\left( U_{i}^{j} \right)}},
	\end{equation}
	
	\noindent
	where $\mu_{k_{P{(U_{i})}}}\left( U_{i}^{j} \right)$ is the membership degree of $U_{i}^{j}$ to set $k_{P{(U_{i})}}$, and $P\left( U_{i} \right) = \left\{ U_{i}^{1},U_{i}^{2},\cdots,U_{i}^{j} \right\}$.
	
	\begin{figure}[htb]%
		\centering
		\includegraphics[width=1\linewidth]{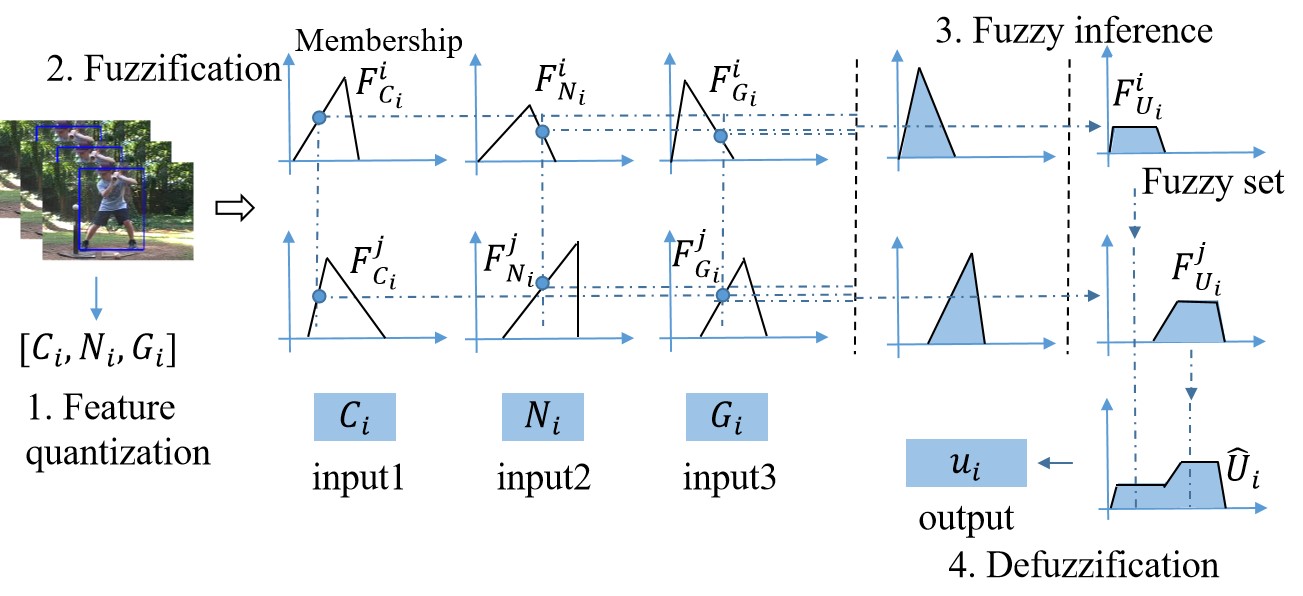}
		\caption{ Illustration of the FCM, which consists of four components: feature quantification, fuzzification, fuzzy inference and defuzzification.}
		\label{fig:fig5}
	\end{figure}
	
	Figure \ref{fig:fig5} shows the overall framework of FCM. First, the corresponding feature vectors are extracted from the action detection and are fuzzified into membership degrees of different sets in the respective universe. Then, fuzzy logic is employed to aggregate the active fuzzy rules. Finally, the effectiveness of frames is obtained through defuzzification.
	
	The FCM evaluates the effectiveness of each frame by incorporating multiple perspectives. This aids in the accurate localization of the detection results with cognitive abnormalities and facilitates the subsequent stages of further detection.
	
	\subsection{Fuzzy cognitive update strategy (FCS)}
	
	The detection performance for each frame in a video sequence depends on the effectiveness of the surrounding action frames. The poor effectiveness of the surrounding action frames often leads to noise interference in detecting the target frame. Therefore, frames are divided into high and low--levels based on their effectiveness. Frames with high--level cognition are considered reliable predictions, whereas the remaining frames require further detection as follows:
	
	\begin{equation}
		\left. D_{A}\leftarrow{\bigcup\left\{ I_{i} \middle| {u_{i} \geq \delta} \right\}},{~D}_{N}\leftarrow I - D_{A} \right..
	\end{equation}
	
	\noindent
	where $D_{A}$ and $D_{N}$ are sets of frames with high and low--level cognition, respectively. $\delta$ is the score threshold.
	
	Then, a storage space is constructed to store correlated sequences composed of frames with high--level cognition, and the features of the correlated sequences are combined to re--detect frames with low--level cognition. Subsequently, the results before and after re--detection are dynamically updated based on fuzzy logic, improving action detection performance. Specifically, the FCS includes three main parts: constructing correlated sequences, re--detecting frames with low--level cognition, and updating frames with low--level cognition.
	
	\noindent
	\textbf{Constructing correlated sequences.} Frames with high--level cognition are employed to construct correlated sequences in re--detection processes. As action detection continues, the size of the correlated sequences gradually increases. The matching of the frame with low--level cognition and all the frames with high--level cognition significantly affects the detection speed. Therefore, for a specific frame with low--level cognition $I_{i}$, only video frames within a certain neighboring range are employed to construct the correlated sequence $T_{i}$.
	
	\begin{equation}
		\left. T_{i}\leftarrow{\bigcup\left\{ I_{j} \middle| I_{j} \in D_{A} \right.},j \in (i - \lambda,i + \lambda)\} \right.,
	\end{equation}
	
	\noindent
	where $\lambda$ is a hyperparameter used to determine the number of correlated sequences around frames with low--level cognition.
	
	\noindent
	\textbf{Re--detecting frames with low--level cognition.} The correlated sequences provide rich contextual clues to determine the category of the current frame. It also contains noise that affects the determination of subsequent frame categories. Therefore, the core idea of re--detection is to simulate local evolution by dynamically aggregating the features of frames with low--level cognition and their correlated sequences.
	
	Specifically, we extract features based on the backbone network in \cite{duan2022revisiting}. For the extracted feature $f_{j}$ of the j--th frame image in the correlated sequence $T_{i}$, $f_{j}$ is converted into the key and value spaces, where the former is responsible for comparing similarities, and the latter can be utilized for feature aggregation. It is expressed as:
	
	\begin{equation}
		f_{j}^{k} = \Phi^{k}\left( f_{j} \right),~~f_{j}^{v} = \Phi^{v}\left( f_{j} \right),
	\end{equation}
	
	\noindent
	where $\Phi^{k}$ and $\Phi^{v}$ denote the key mapping convolutional layer and value mapping convolutional layer in the re--detection network, respectively.
	
	Then, the cosine similarity is calculated to measure the similarity between the feature of the correlated sequence $f_{j}^{k}$ and the feature of the frame with low--level cognition $f_{i}^{k}$, as follows:
	
	\begin{equation}
		\varepsilon_{j,i}=\frac{f_j^k\cdot f_i^k}{\left\|f_j^k\right\|\cdot\left\|f_i^k\right\|},
	\end{equation}
	
	\noindent
	where $\varepsilon_{j,i}$ refers to the similarity between the j--th frame of the correlated sequence  $f_{j}^{k}$ and $f_{i}^{k}$. The similarity set $\left\lbrack \varepsilon_{1,i},\cdots,\varepsilon_{j,i} \right\rbrack$ is obtained from above, $j \in (1,2,\cdots,z)$ is the number of frames in the correlated sequence, and softmax is employed to normalize and obtain the attention mask $\left\{ \hat{\varepsilon} \right\}$. The feature values in the correlated sequence are aggregated to obtain features with high--level cognition $\tilde{f}_{i}$:
	
	\begin{equation}
		\tilde{f}_{i}=\sum_{j=1}^{z}\hat{\varepsilon}_{j,i}\cdot f_{j}^{v}.
	\end{equation}
	
	Then, the features of the frames with low--level cognition $f_{i}^{v}$ and high--level cognition $\tilde{f}_{i}$ are employed to obtain the predicted classification score $H_{i}$ after re--detection.
	
	\begin{equation}
		H_i=\Omega(\tilde{f}_i,f_i^v),
	\end{equation}
	
	\noindent
	where $\Omega(\cdot,\cdot)$ is the classifier of the re--detection network, $H_{i} = \left( C^{K_{1}},\cdots,C^{K_{h}} \right)$ indicates the set of classification scores after re--detection, and $C^{K_{h}}$ represents the predicted score of $K_{h}$. The infinite norm of the classification is considered the confidence ${\hat{C}}_{i}$ of the re--detection.
	
	\begin{equation}
		\hat{C}_i=\|H_i\|_\infty . 
	\end{equation}
	
	\noindent
	\textbf{Updating frames with low--level cognition.} From the above steps, confidence ${\hat{C}}_{i}$ after re--detection is obtained. However, updating the results is unreliable when only confidence is employed after re--detection. Therefore, it is necessary to replace $C_{i}$ with ${\hat{C}}_{i}$ as the input for fuzzy inference and use Equations (5)--(8) to obtain the effectiveness ${\hat{u}}_{i}$ after the re--detection. Then, an updater is designed to dynamically update the results based on the effectiveness before and after re--detection, as follows:
	
	\begin{equation}
		u_{opt}^{i}=\max\left(\hat{u}_{i},u_{i}+\tau\right).
	\end{equation}
	
	\noindent
	where $u_{i}$ is the effectiveness of the initial frame and ${\hat{u}}_{i}$ refers to the effectiveness of the frame after re--detection. $\tau$ denotes a predefined threshold. Because the initial frame is the foundation of the detection process, the result's update occurs only when the combined effectiveness of the initial frame and the predefined threshold is lower than the effectiveness obtained after re--detection.
	
	\begin{figure}[htb]%
		\centering
		\includegraphics[width=0.95\linewidth]{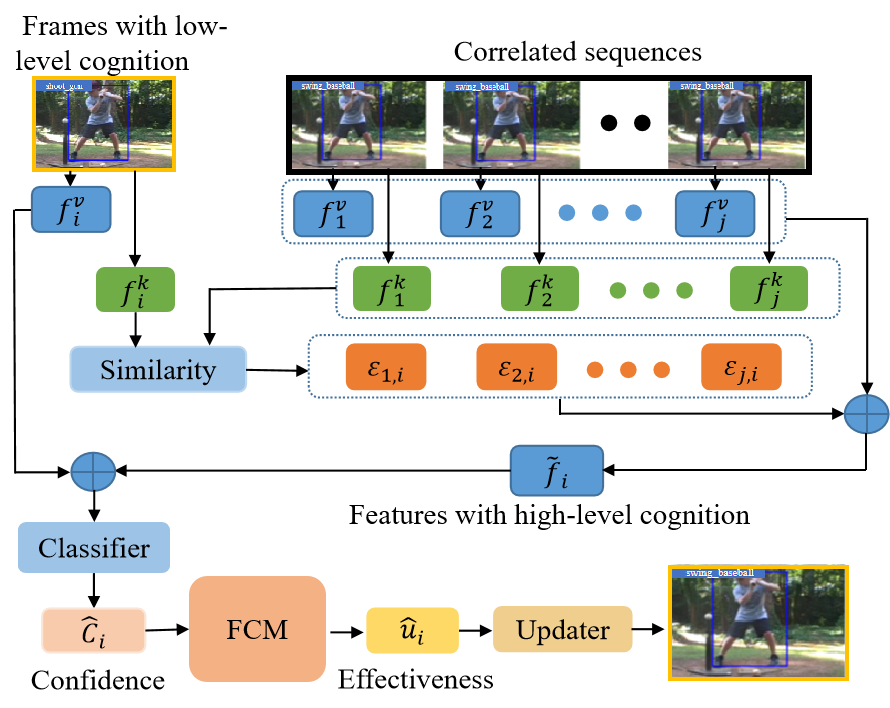}
		\caption{Illustration of the FCS. FCS measures the similarity between features in correlated sequences and frames with low--level cognitive abilities, and fuses the features to obtain features with high--level cognitive abilities.}
		\label{fig:fig6}
	\end{figure}
	
	Figure \ref{fig:fig6} shows the overall process of the FCS. First, the correlated sequences of frames with high--level cognition are constructed for re--detection. Then, the similarity between the features in the correlated sequence and frames with low--level cognition is measured, and the features are fused to obtain features with high--level cognition. The features of frames with low--level cognition and correlated sequences are jointly considered for re--detection. Finally, fuzzy logic is employed to dynamically update the detection results with low--level cognition before and after re--detection.

	\section{Experimental results and analysis}
	
	To evaluate the superiority of Cefdet for action detection, experiments are conducted using JHMDB \cite{jhuang2013towards}, UCF101--24 \cite{soomro2012ucf101} and AVA2.2 \cite{gu2018ava}. The following introduces these datasets and describes their implementation details. Then, the experimental results of Cefdet are quantitatively and qualitatively discussed to demonstrate its effectiveness.
	
	\subsection{Datasets}
	
	\textbf{JHMDB} \cite{jhuang2013towards} is a benchmark dataset for action detection. It consists of video composed of 928 temporary clips from 21 different action categories. The dataset involves fine-grained actions and subtle temporal cues, requiring precise temporal localization for accurate detection.
	
	\noindent
	\textbf{UCF101--24} \cite{soomro2012ucf101} is derived from the UCF101 dataset and focuses on 24 action categories consisting of 3207 videos representing various human activities such as walking, jogging, basketball, and dancing. These videos exhibit variations in viewpoint, scale, background clutter, and lighting conditions, posing challenges for accurate action detection.
	
	\noindent
	\textbf{AVA2.2} \cite{gu2018ava} is a large video dataset for action recognition and understanding, containing 430 15--minute video clips. The dataset covers up to 80 different categories of labels, such as "handshake", "jump", "kiss", etc., all of which are finely annotated. Following standard settings, this study reports frame mAPs for 60 out of 80 categories, with an IoU threshold of 0.5.
	
	\subsection{Implementation details}
	
	\textbf{Object detector:} The experiment uses the character bounding boxes detected in \cite{kopuklu2019you} for inference. The target detector adopts a Faster RCNN \cite{ren2016faster} in the ResNet--50--FPN backbone. The model is pretrained on ImageNet \cite{deng2009imagenet} and fine-tuned on MSCOCO \cite{lin2014microsoft}. 
	
	\noindent
	\textbf{Backbone:} SlowFast \cite{feichtenhofer2019slowfast} is employed as the video backbone. The experiment is instantiated using SlowFast and ResNet--50 pretrained on Kinetics700 \cite{carreira2017quo}. 
	
	\noindent
	\textbf{Training and evaluation:} The model is trained for 7k iterations on the JHMDB dataset, where the first 700 iterations are for linear preheating. SGD is used as the optimizer with a batch size of 8. Similarly, there are 50k iterations of training on the UCF101--24 dataset, with linear preheating applied in the first 1k iterations. The learning rate is 0.0002, which is 10 times less in iterations of 25k and 35k. The hyperparameters $\mu_{1}$ and $\mu_{2}$ are 0.6 and 0.2, respectively.
	
	\subsection{Quantitative analysis}
	
	To verify the effectiveness of Cefdet in action detection, we compare it with state--of--the--art (SOTA) methods using the JHMDB and UCF101--24 datasets. The frame mean average precision (mAP) with an intersection over union (IoU) threshold of 0.5 is employed as the evaluation metric, and experimental results are shown in Table \ref{tab:tab2} and \ref{tab:tab3}.
	
	\begin{table}
		\caption{Experimental results using the JHMDB dataset based on mAP. Cefdet exhibits excellent performance on both frame mAP and video mAP.}
		\label{tab:tab2}
		\begin{tabular}{lllll}
			\toprule
			Model & Input & f@0.5 & v@0.2 & v@0.5 \\
			\midrule
			MOC(2020)\cite{li2020actions} & V+F & 70.8 & 77.3 & 77.2 \\
			AVA(2018)\cite{gu2018ava} & V+F & 73.3 & -- & 78.6 \\
			PCSC(2019)\cite{su2019improving} & V+F & 74.8 & 82.6 & 82.2 \\
			HISAN(2019)\cite{pramono2019hierarchical} & V+F & 76.7 & 85.9 & 84.0 \\
			ACRN(2018)\cite{sun2018actor} & V+F & 77.9 & -- & 80.0 \\
			CA--RCNN(2020)\cite{wu2020context} & V & 79.2 & -- & -- \\
			WOO(2021)\cite{chen2021watch} & V & 80.5 & -- & -- \\
			TubeR(2022)\cite{zhao2022tuber} & V+F & -- & 87.4 & 82.3 \\
			SE--STAD(2023)\cite{sui2023simple} & V & 82.5 & -- & -- \\
			MCA--SVMM(2023)\cite{zhao2023modulation} & V & 74.9 & 82.1 & 81.8 \\
			HIT(2023)\cite{faure2023holistic} & V & 82.9 & 88.5 & 87.2 \\
			\midrule
			\textbf{Ours} & V & \textbf{84.0} & \textbf{90.4} & \textbf{89.3} \\
			\bottomrule
		\end{tabular}
	\end{table}
	
	\begin{table}
		\caption{Experimental results on the UCF101--24 dataset based on mAP. Cefdet surpasses SOTA methods.}
		\label{tab:tab3}
		\begin{tabular}{lllll}
			\toprule
			Model & Input & f@0.5 & v@0.2 & v@0.5 \\
			\midrule
			HISAN(2019)\cite{pramono2019hierarchical} & V+F & 73.7 & 80.4 & 49.5 \\
			MOC(2020)\cite{li2020actions} & V+F & 78.0 & 82.8 & 53.8 \\
			AVA(2018)\cite{gu2018ava} & V & 76.3 & -- & -- \\
			AIA(2020)\cite{tang2020asynchronous} & V & 78.8 & -- & -- \\
			PCSC(2019)\cite{su2019improving} & V+F & 79.2 & 84.3 & 61.0 \\
			TubeR(2022)\cite{zhao2022tuber} & V+F & 83.2 & 83.3 & 58.4 \\
			ACAR(2021)\cite{pan2021actor} & V & 84.3 & -- & -- \\
			CycleACR(2023)\cite{chen2023cycleacr} & V & 84.7 & -- & -- \\
			MCA--SVMM(2023)\cite{zhao2023modulation} & V & 79.3 & 83.4 & 54.6 \\
			HIT(2023)\cite{faure2023holistic} & V & 84.8 & 88.8 & 74.3 \\
			\midrule
			\textbf{Ours} & V & \textbf{85.1} & \textbf{89.3} & \textbf{75.1} \\
			\bottomrule
		\end{tabular}
	\end{table}
	
	\begin{table}
		\caption{Experimental results on the AVA2.2 dataset based on mAP. Cefdet has comparable results compared to the SOTA methods.}
		\label{tab:tab7}
		\begin{tabular}{lll}
			\toprule
			Model & Pretrain & mAP \\
			\midrule
			SlowFast, R-101+NL(2019)\cite{feichtenhofer2019slowfast} & K600 & 29.0 \\
			X3D-L(2020)\cite{feichtenhofer2020x3d} & K600 & 29.4 \\
			AIA(2020)\cite{tang2020asynchronous} & K700 & 32.3 \\
			Object Transformer(2021)\cite{wu2021towards} & K600 & 31.0 \\
			Beyond Short Clips(2021)\cite{yang2021beyond} & K700 & 31.6 \\
			TubeR(2022)\cite{zhao2022tuber} & IG + 400 & \textbf{33.6} \\
			HIT(2023)\cite{faure2023holistic} & K700 & 32.4 \\
			\midrule
			\textbf{Ours} & K700 & 32.7 \\
			\bottomrule
		\end{tabular}
	\end{table}
	
	The mAP of SOTA reaches 82.9\% and 84.8\% for the JHMDB and UCF101--24 datasets, respectively. Although the SOTA algorithm exhibits advanced performance on the JHMDB and UCF101--24 datasets, its detection accuracy for high--similarity actions remains poor. Moreover, it is difficult for a detector to explore the relationship between continuous actions, resulting in detection results that do not conform to human action norms. Repairing these detection results with cognitive abnormalities is a challenge in action detection because of the lack of effective judgment of the detection results.
	
	However, Cefdet achieves mAP values of 84.0\% and 85.0\% on the JHMDB and UCF101--24 datasets, respectively, which provides gains of 1.1\% and 0.2\%, respectively, compared with SOTA. Cefdet utilizes fuzzy inference to simulate a cognition--based detection process, effectively determining the effectiveness of frames and accurately locating frames with cognitive abnormalities. Furthermore, each frame is divided into frames with high and low--level cognition based on their effectiveness. The features of the frames with high--level cognition are refined to assist in the re--detection of frames with low--level cognition. Then, fuzzy logic is employed to dynamically update the detection results with low--level cognition before and after re--detection. This reduces the impact of noise and misjudgment, effectively repairs the detection results with cognitive abnormalities.
	
	To further validate the generalization ability of Cefdet, this study conducts experiments on the most challenging fine--grained action detection dataset (AVA2.2), as shown in Table \ref{tab:tab7}. Cefdet achieves the mAP value of 32.7\%, demonstrating competitive results. TubeR \cite{zhao2022tuber} achieves the best results thanks to using a backbone pretrained on the IG+K400 dataset. Within the same pretrained backbone, Cefdet consistently reports the excellent performance. Overall, the experimental results on three benchmark datasets demonstrate the generalization ability of Cefdet.
	
	Table \ref{tab:tab4} indicate that the detection accuracy of action categories such as “shoot\_ball”, “walk”, and “stand” is poor on the JHMDB dataset. Existing methods are difficult to distinguish similar actions. However, Cefdet exhibits excellent performance in these action categories, with improvements of 12.2\%, 10.5\%, and 7.2\%, respectively, compared with the SOTA algorithm. This improvement is attributed to evaluating the effectiveness of the fuzzy inference system. It simulates a cognition--based detection process and accurately identifies the location of the detection results with cognitive abnormalities. Subsequently, the detection results are divided into frames with high and low--level cognition based on their effectiveness. Frames with high--level cognition guide frames with low--level for re--detection, which effectively repairs detection results with cognitive abnormalities.
	
	\begin{table}
		\caption{Experimental results of representative actions in the JHMDB dataset (mAP).}
		\label{tab:tab4}
		\begin{tabular}{llll}
			\toprule
			Action & HIT(\%) & Ours(\%) & Gap(\%) \\
			\midrule
			Shoot\_ball & 71.2 & 83.4 & +12.2 \\
			Sit & 52.3 & 56.5 & +4.2 \\
			Stand & 26.1 & 33.3 & +7.2 \\
			Walk & 78.6 & 89.1 & +10.5 \\
			Climb\_stairs & 89.2 & 91.3 & +2.1 \\
			Throw & 75.5 & 78.1 & +2.6 \\
			\bottomrule
		\end{tabular}
	\end{table}
	
	The above experimental results demonstrate that Cefdet achieves SOTA performance on the JHMDB and UCF101--24 datasets, and show competitive performance on the AVA2.2 dataset. This fully reflects Cefdet’s superiority and robustness.
	
	\subsection{Ablation studies}
	
	In this section, ablation experiments are conducted on JHMDB to investigate the effectiveness of FCM and FCS. The frame mAP with an IoU threshold of 0.5 is employed as an evaluation metric, and the experimental results are shown in Table \ref{tab:tab5} and \ref{tab:tab6}.
	
	
	\begin{table}
		\caption{The ablation experiment in the JHMDB dataset.}
		\label{tab:tab5}
		\begin{tabular}{lll}
			\toprule
			FCM & FCS & mAP \\
			\midrule
			&   & 82.9 \\
			\Checkmark &  & 83.8 \\
			& \Checkmark & 83.2 \\
			\Checkmark & \Checkmark & \textbf{84.0} \\
			\bottomrule
		\end{tabular}
	\end{table}
	
	\begin{table}
		\caption{Results of fuzzy cognitive update under different thresholds using confidence and effectiveness as evaluation criteria in the JHMDB dataset (mAP).}
		\label{tab:tab6}
		\begin{tabular}{lllllll}
			\toprule
			Threshold & 0.1 & 0.2 & 0.3 & 0.35 & 0.4 & 0.5 \\
			\midrule
			Confidence & 83.0  & 83.1 & \textbf{83.2} & 83.2 & 83.1 & 83.1 \\
			Effectiveness & 83.4 & 83.6 & 83.8 & \textbf{84.0} & 83.9 & 83.8 \\
			\bottomrule
		\end{tabular}
	\end{table}
	
	Notably, when FCM is removed, the confidence score for action detection is selected as the evaluation criterion. The obtained mAP is 83.2\%, which is a reduction of 0.8\%, indicating that effectiveness has a more significant effect than confidence. Table \ref{tab:tab6} presents the results of using confidence and effectiveness as the evaluation criteria. The effectiveness as the evaluation criterion are consistently higher than the confidence under different thresholds. The highest mAP of 84.0 is achieved at a threshold of 0.35, indicating that cognition--based detection achieves accurate localization of the positions of the detection results with cognitive abnormalities.
	
	The experimental results are updated based on confidence before and after re--detection after removing the FCS. The result is a 0.2\% decrease, demonstrating that using only confidence scores to update the experimental results is unreasonable. The FCS has demonstrated superior capabilities, effectively improving the accuracy of detecting cognitive abnormalities. 
	
	The results demonstrate that both the FCM and FCS effectively enhance the accuracy of action detection in complex scenarios.

	\section{Conclusion and future work}
	
	\begin{sloppypar}
		This study proposes Cefdet, which introduces the concept of “cognition--based detection” into action detection to simulate human cognition. First, the FCM is designed to evaluate the effectiveness of frames. It extracts various action features from video sequences and inputs them into fuzzy inference. The cognition--based detection process is simulated using fuzzy inference to obtain effectiveness, which accurately locates the position of action frames with cognitive abnormalities. Then, the FCS is proposed based on the FCM to update the results dynamically. It constructs correlated sequences by frames with high--level cognition and performs a weighted fusion of features by fuzzy inference for re--detection. Subsequently, the results before and after re--detection are dynamically updated, which effectively repairs the detection results with cognitive abnormalities. Experiments on public datasets prove that Cefdet achieves superior performance and promotes the further application of fuzzy inference in action detection.
		
		In future work, our objective will be to incorporate interval type--2 fuzzy set theory into developing a robust and real--time action detection framework and further facilitate the application of fuzzy theory in action detection.
	\end{sloppypar}
	
	\clearpage

	\begin{acks}
		\begin{sloppypar}
			This work was supported in part by the Key Project of Higher Education Teaching Reform Research in Hunan Province (No. HNJG-20230227) and the Natural Science Foundation of China (No. 62207012).
		\end{sloppypar}
	\end{acks}

	\bibliographystyle{ACM-Reference-Format}
	\balance
	\bibliography{acmart}


\begin{thebibliography}{47}


\ifx \showCODEN    \undefined \def \showCODEN     #1{\unskip}     \fi
\ifx \showDOI      \undefined \def \showDOI       #1{#1}\fi
\ifx \showISBNx    \undefined \def \showISBNx     #1{\unskip}     \fi
\ifx \showISBNxiii \undefined \def \showISBNxiii  #1{\unskip}     \fi
\ifx \showISSN     \undefined \def \showISSN      #1{\unskip}     \fi
\ifx \showLCCN     \undefined \def \showLCCN      #1{\unskip}     \fi
\ifx \shownote     \undefined \def \shownote      #1{#1}          \fi
\ifx \showarticletitle \undefined \def \showarticletitle #1{#1}   \fi
\ifx \showURL      \undefined \def \showURL       {\relax}        \fi
\providecommand\bibfield[2]{#2}
\providecommand\bibinfo[2]{#2}
\providecommand\natexlab[1]{#1}
\providecommand\showeprint[2][]{arXiv:#2}

\bibitem[Bao et~al\mbox{.}(2022)]%
        {bao2022hierarchical}
\bibfield{author}{\bibinfo{person}{Qianyue Bao}, \bibinfo{person}{Fang Liu}, \bibinfo{person}{Yang Liu}, \bibinfo{person}{Licheng Jiao}, \bibinfo{person}{Xu Liu}, {and} \bibinfo{person}{Lingling Li}.} \bibinfo{year}{2022}\natexlab{}.
\newblock \showarticletitle{Hierarchical scene normality-binding modeling for anomaly detection in surveillance videos}. In \bibinfo{booktitle}{\emph{Proceedings of the 30th ACM international conference on multimedia}}. \bibinfo{pages}{6103--6112}.
\newblock


\bibitem[Cao et~al\mbox{.}(2020)]%
        {cao2020multiobjective}
\bibfield{author}{\bibinfo{person}{Bin Cao}, \bibinfo{person}{Jianwei Zhao}, \bibinfo{person}{Zhihan Lv}, \bibinfo{person}{Yu Gu}, \bibinfo{person}{Peng Yang}, {and} \bibinfo{person}{Saman~K Halgamuge}.} \bibinfo{year}{2020}\natexlab{}.
\newblock \showarticletitle{Multiobjective evolution of fuzzy rough neural network via distributed parallelism for stock prediction}.
\newblock \bibinfo{journal}{\emph{IEEE Transactions on Fuzzy Systems}} \bibinfo{volume}{28}, \bibinfo{number}{5} (\bibinfo{year}{2020}), \bibinfo{pages}{939--952}.
\newblock


\bibitem[Carreira and Zisserman(2017)]%
        {carreira2017quo}
\bibfield{author}{\bibinfo{person}{Joao Carreira} {and} \bibinfo{person}{Andrew Zisserman}.} \bibinfo{year}{2017}\natexlab{}.
\newblock \showarticletitle{Quo vadis, action recognition? a new model and the kinetics dataset}. In \bibinfo{booktitle}{\emph{proceedings of the IEEE Conference on Computer Vision and Pattern Recognition}}. \bibinfo{pages}{6299--6308}.
\newblock


\bibitem[Chen et~al\mbox{.}(2023)]%
        {chen2023cycleacr}
\bibfield{author}{\bibinfo{person}{Lei Chen}, \bibinfo{person}{Zhan Tong}, \bibinfo{person}{Yibing Song}, \bibinfo{person}{Gangshan Wu}, {and} \bibinfo{person}{Limin Wang}.} \bibinfo{year}{2023}\natexlab{}.
\newblock \showarticletitle{Cycleacr: Cycle modeling of actor-context relations for video action detection}.
\newblock \bibinfo{journal}{\emph{arXiv preprint arXiv:2303.16118}} (\bibinfo{year}{2023}).
\newblock


\bibitem[Chen et~al\mbox{.}(2021a)]%
        {chen2021watch}
\bibfield{author}{\bibinfo{person}{Shoufa Chen}, \bibinfo{person}{Peize Sun}, \bibinfo{person}{Enze Xie}, \bibinfo{person}{Chongjian Ge}, \bibinfo{person}{Jiannan Wu}, \bibinfo{person}{Lan Ma}, \bibinfo{person}{Jiajun Shen}, {and} \bibinfo{person}{Ping Luo}.} \bibinfo{year}{2021}\natexlab{a}.
\newblock \showarticletitle{Watch only once: An end-to-end video action detection framework}. In \bibinfo{booktitle}{\emph{Proceedings of the IEEE/CVF International Conference on Computer Vision}}. \bibinfo{pages}{8178--8187}.
\newblock


\bibitem[Chen et~al\mbox{.}(2021b)]%
        {chen2021learning}
\bibfield{author}{\bibinfo{person}{Tailin Chen}, \bibinfo{person}{Desen Zhou}, \bibinfo{person}{Jian Wang}, \bibinfo{person}{Shidong Wang}, \bibinfo{person}{Yu Guan}, \bibinfo{person}{Xuming He}, {and} \bibinfo{person}{Errui Ding}.} \bibinfo{year}{2021}\natexlab{b}.
\newblock \showarticletitle{Learning multi-granular spatio-temporal graph network for skeleton-based action recognition}. In \bibinfo{booktitle}{\emph{Proceedings of the 29th ACM international conference on multimedia}}. \bibinfo{pages}{4334--4342}.
\newblock


\bibitem[Cheng et~al\mbox{.}(2023)]%
        {cheng2023sample}
\bibfield{author}{\bibinfo{person}{Harry Cheng}, \bibinfo{person}{Yangyang Guo}, \bibinfo{person}{Liqiang Nie}, \bibinfo{person}{Zhiyong Cheng}, {and} \bibinfo{person}{Mohan Kankanhalli}.} \bibinfo{year}{2023}\natexlab{}.
\newblock \showarticletitle{Sample less, learn more: Efficient action recognition via frame feature restoration}. In \bibinfo{booktitle}{\emph{Proceedings of the 31st ACM International Conference on Multimedia}}. \bibinfo{pages}{7101--7110}.
\newblock


\bibitem[Deng et~al\mbox{.}(2009)]%
        {deng2009imagenet}
\bibfield{author}{\bibinfo{person}{Jia Deng}, \bibinfo{person}{Wei Dong}, \bibinfo{person}{Richard Socher}, \bibinfo{person}{Li-Jia Li}, \bibinfo{person}{Kai Li}, {and} \bibinfo{person}{Li Fei-Fei}.} \bibinfo{year}{2009}\natexlab{}.
\newblock \showarticletitle{Imagenet: A large-scale hierarchical image database}. In \bibinfo{booktitle}{\emph{2009 IEEE conference on computer vision and pattern recognition}}. Ieee, \bibinfo{pages}{248--255}.
\newblock


\bibitem[Donahue et~al\mbox{.}(2015)]%
        {donahue2015long}
\bibfield{author}{\bibinfo{person}{Jeffrey Donahue}, \bibinfo{person}{Lisa Anne~Hendricks}, \bibinfo{person}{Sergio Guadarrama}, \bibinfo{person}{Marcus Rohrbach}, \bibinfo{person}{Subhashini Venugopalan}, \bibinfo{person}{Kate Saenko}, {and} \bibinfo{person}{Trevor Darrell}.} \bibinfo{year}{2015}\natexlab{}.
\newblock \showarticletitle{Long-term recurrent convolutional networks for visual recognition and description}. In \bibinfo{booktitle}{\emph{Proceedings of the IEEE conference on computer vision and pattern recognition}}. \bibinfo{pages}{2625--2634}.
\newblock


\bibitem[Duan et~al\mbox{.}(2022)]%
        {duan2022revisiting}
\bibfield{author}{\bibinfo{person}{Haodong Duan}, \bibinfo{person}{Yue Zhao}, \bibinfo{person}{Kai Chen}, \bibinfo{person}{Dahua Lin}, {and} \bibinfo{person}{Bo Dai}.} \bibinfo{year}{2022}\natexlab{}.
\newblock \showarticletitle{Revisiting skeleton-based action recognition}. In \bibinfo{booktitle}{\emph{Proceedings of the IEEE/CVF conference on computer vision and pattern recognition}}. \bibinfo{pages}{2969--2978}.
\newblock


\bibitem[Fan and Kankanhalli(2021)]%
        {fan2021motion}
\bibfield{author}{\bibinfo{person}{Hehe Fan} {and} \bibinfo{person}{Mohan Kankanhalli}.} \bibinfo{year}{2021}\natexlab{}.
\newblock \showarticletitle{Motion= video-content: Towards unsupervised learning of motion representation from videos}. In \bibinfo{booktitle}{\emph{Proceedings of the 3rd ACM International Conference on Multimedia in Asia}}. \bibinfo{pages}{1--7}.
\newblock


\bibitem[Faure et~al\mbox{.}(2023)]%
        {faure2023holistic}
\bibfield{author}{\bibinfo{person}{Gueter~Josmy Faure}, \bibinfo{person}{Min-Hung Chen}, {and} \bibinfo{person}{Shang-Hong Lai}.} \bibinfo{year}{2023}\natexlab{}.
\newblock \showarticletitle{Holistic interaction transformer network for action detection}. In \bibinfo{booktitle}{\emph{Proceedings of the IEEE/CVF Winter Conference on Applications of Computer Vision}}. \bibinfo{pages}{3340--3350}.
\newblock


\bibitem[Feichtenhofer(2020)]%
        {feichtenhofer2020x3d}
\bibfield{author}{\bibinfo{person}{Christoph Feichtenhofer}.} \bibinfo{year}{2020}\natexlab{}.
\newblock \showarticletitle{X3d: Expanding architectures for efficient video recognition}. In \bibinfo{booktitle}{\emph{Proceedings of the IEEE/CVF conference on computer vision and pattern recognition}}. \bibinfo{pages}{203--213}.
\newblock


\bibitem[Feichtenhofer et~al\mbox{.}(2019)]%
        {feichtenhofer2019slowfast}
\bibfield{author}{\bibinfo{person}{Christoph Feichtenhofer}, \bibinfo{person}{Haoqi Fan}, \bibinfo{person}{Jitendra Malik}, {and} \bibinfo{person}{Kaiming He}.} \bibinfo{year}{2019}\natexlab{}.
\newblock \showarticletitle{Slowfast networks for video recognition}. In \bibinfo{booktitle}{\emph{Proceedings of the IEEE/CVF international conference on computer vision}}. \bibinfo{pages}{6202--6211}.
\newblock


\bibitem[Gu et~al\mbox{.}(2018)]%
        {gu2018ava}
\bibfield{author}{\bibinfo{person}{Chunhui Gu}, \bibinfo{person}{Chen Sun}, \bibinfo{person}{David~A Ross}, \bibinfo{person}{Carl Vondrick}, \bibinfo{person}{Caroline Pantofaru}, \bibinfo{person}{Yeqing Li}, \bibinfo{person}{Sudheendra Vijayanarasimhan}, \bibinfo{person}{George Toderici}, \bibinfo{person}{Susanna Ricco}, \bibinfo{person}{Rahul Sukthankar}, {et~al\mbox{.}}} \bibinfo{year}{2018}\natexlab{}.
\newblock \showarticletitle{Ava: A video dataset of spatio-temporally localized atomic visual actions}. In \bibinfo{booktitle}{\emph{Proceedings of the IEEE conference on computer vision and pattern recognition}}. \bibinfo{pages}{6047--6056}.
\newblock


\bibitem[Guan et~al\mbox{.}(2023)]%
        {guan2023egocentric}
\bibfield{author}{\bibinfo{person}{Weili Guan}, \bibinfo{person}{Xuemeng Song}, \bibinfo{person}{Kejie Wang}, \bibinfo{person}{Haokun Wen}, \bibinfo{person}{Hongda Ni}, \bibinfo{person}{Yaowei Wang}, {and} \bibinfo{person}{Xiaojun Chang}.} \bibinfo{year}{2023}\natexlab{}.
\newblock \showarticletitle{Egocentric early action prediction via multimodal transformer-based dual action prediction}.
\newblock \bibinfo{journal}{\emph{IEEE Transactions on Circuits and Systems for Video Technology}} \bibinfo{volume}{33}, \bibinfo{number}{9} (\bibinfo{year}{2023}), \bibinfo{pages}{4472--4483}.
\newblock


\bibitem[Jhuang et~al\mbox{.}(2013)]%
        {jhuang2013towards}
\bibfield{author}{\bibinfo{person}{Hueihan Jhuang}, \bibinfo{person}{Juergen Gall}, \bibinfo{person}{Silvia Zuffi}, \bibinfo{person}{Cordelia Schmid}, {and} \bibinfo{person}{Michael~J Black}.} \bibinfo{year}{2013}\natexlab{}.
\newblock \showarticletitle{Towards understanding action recognition}. In \bibinfo{booktitle}{\emph{Proceedings of the IEEE international conference on computer vision}}. \bibinfo{pages}{3192--3199}.
\newblock


\bibitem[Kim et~al\mbox{.}(2017)]%
        {kim2017design}
\bibfield{author}{\bibinfo{person}{Eun-Hu Kim}, \bibinfo{person}{Sung-Kwun Oh}, {and} \bibinfo{person}{Witold Pedrycz}.} \bibinfo{year}{2017}\natexlab{}.
\newblock \showarticletitle{Design of reinforced interval type-2 fuzzy c-means-based fuzzy classifier}.
\newblock \bibinfo{journal}{\emph{IEEE Transactions on Fuzzy Systems}} \bibinfo{volume}{26}, \bibinfo{number}{5} (\bibinfo{year}{2017}), \bibinfo{pages}{3054--3068}.
\newblock


\bibitem[K{\"o}p{\"u}kl{\"u} et~al\mbox{.}(2019)]%
        {kopuklu2019you}
\bibfield{author}{\bibinfo{person}{Okan K{\"o}p{\"u}kl{\"u}}, \bibinfo{person}{Xiangyu Wei}, {and} \bibinfo{person}{Gerhard Rigoll}.} \bibinfo{year}{2019}\natexlab{}.
\newblock \showarticletitle{You only watch once: A unified cnn architecture for real-time spatiotemporal action localization}.
\newblock \bibinfo{journal}{\emph{arXiv preprint arXiv:1911.06644}} (\bibinfo{year}{2019}).
\newblock


\bibitem[Li et~al\mbox{.}(2017)]%
        {li2017optimal}
\bibfield{author}{\bibinfo{person}{Hongyi Li}, \bibinfo{person}{Jiahui Wang}, \bibinfo{person}{Ligang Wu}, \bibinfo{person}{Hak-Keung Lam}, {and} \bibinfo{person}{Yabin Gao}.} \bibinfo{year}{2017}\natexlab{}.
\newblock \showarticletitle{Optimal guaranteed cost sliding-mode control of interval type-2 fuzzy time-delay systems}.
\newblock \bibinfo{journal}{\emph{IEEE Transactions on Fuzzy Systems}} \bibinfo{volume}{26}, \bibinfo{number}{1} (\bibinfo{year}{2017}), \bibinfo{pages}{246--257}.
\newblock


\bibitem[Li et~al\mbox{.}(2020)]%
        {li2020actions}
\bibfield{author}{\bibinfo{person}{Yixuan Li}, \bibinfo{person}{Zixu Wang}, \bibinfo{person}{Limin Wang}, {and} \bibinfo{person}{Gangshan Wu}.} \bibinfo{year}{2020}\natexlab{}.
\newblock \showarticletitle{Actions as moving points}. In \bibinfo{booktitle}{\emph{Computer Vision--ECCV 2020: 16th European Conference, Glasgow, UK, August 23--28, 2020, Proceedings, Part XVI 16}}. Springer, \bibinfo{pages}{68--84}.
\newblock


\bibitem[Lin et~al\mbox{.}(2014)]%
        {lin2014microsoft}
\bibfield{author}{\bibinfo{person}{Tsung-Yi Lin}, \bibinfo{person}{Michael Maire}, \bibinfo{person}{Serge Belongie}, \bibinfo{person}{James Hays}, \bibinfo{person}{Pietro Perona}, \bibinfo{person}{Deva Ramanan}, \bibinfo{person}{Piotr Doll{\'a}r}, {and} \bibinfo{person}{C~Lawrence Zitnick}.} \bibinfo{year}{2014}\natexlab{}.
\newblock \showarticletitle{Microsoft coco: Common objects in context}. In \bibinfo{booktitle}{\emph{Computer Vision--ECCV 2014: 13th European Conference, Zurich, Switzerland, September 6-12, 2014, Proceedings, Part V 13}}. Springer, \bibinfo{pages}{740--755}.
\newblock


\bibitem[Liu et~al\mbox{.}(2023)]%
        {liu2023weakly}
\bibfield{author}{\bibinfo{person}{Min Liu}, \bibinfo{person}{Yuan Bian}, \bibinfo{person}{Qing Liu}, \bibinfo{person}{Xueping Wang}, {and} \bibinfo{person}{Yaonan Wang}.} \bibinfo{year}{2023}\natexlab{}.
\newblock \showarticletitle{Weakly supervised tracklet association learning with video labels for person re-identification}.
\newblock \bibinfo{journal}{\emph{IEEE Transactions on Pattern Analysis and Machine Intelligence}} (\bibinfo{year}{2023}), \bibinfo{pages}{3595--3607}.
\newblock


\bibitem[Liu et~al\mbox{.}(2024b)]%
        {liu2024two}
\bibfield{author}{\bibinfo{person}{Min Liu}, \bibinfo{person}{Fei Wang}, \bibinfo{person}{Xueping Wang}, \bibinfo{person}{Yaonan Wang}, {and} \bibinfo{person}{Amit~K Roy-Chowdhury}.} \bibinfo{year}{2024}\natexlab{b}.
\newblock \showarticletitle{A two-stage noise-tolerant paradigm for label corrupted person re-identification}.
\newblock \bibinfo{journal}{\emph{IEEE Transactions on Pattern Analysis and Machine Intelligence}} (\bibinfo{year}{2024}), \bibinfo{pages}{4944--4956}.
\newblock


\bibitem[Liu et~al\mbox{.}(2020)]%
        {liu2020fuzzy}
\bibfield{author}{\bibinfo{person}{Shuai Liu}, \bibinfo{person}{Shuai Wang}, \bibinfo{person}{Xinyu Liu}, \bibinfo{person}{Chin-Teng Lin}, {and} \bibinfo{person}{Zhihan Lv}.} \bibinfo{year}{2020}\natexlab{}.
\newblock \showarticletitle{Fuzzy detection aided real-time and robust visual tracking under complex environments}.
\newblock \bibinfo{journal}{\emph{IEEE Transactions on Fuzzy Systems}} \bibinfo{volume}{29}, \bibinfo{number}{1} (\bibinfo{year}{2020}), \bibinfo{pages}{90--102}.
\newblock


\bibitem[Liu et~al\mbox{.}(2024a)]%
        {liu2024knowledge}
\bibfield{author}{\bibinfo{person}{Yang Liu}, \bibinfo{person}{Fang Liu}, \bibinfo{person}{Licheng Jiao}, \bibinfo{person}{Qianyue Bao}, \bibinfo{person}{Lingling Li}, \bibinfo{person}{Yuwei Guo}, {and} \bibinfo{person}{Puhua Chen}.} \bibinfo{year}{2024}\natexlab{a}.
\newblock \showarticletitle{A knowledge-based hierarchical causal inference network for video action recognition}.
\newblock \bibinfo{journal}{\emph{IEEE Transactions on Multimedia}} (\bibinfo{year}{2024}), \bibinfo{pages}{1--16}.
\newblock


\bibitem[Pan et~al\mbox{.}(2021)]%
        {pan2021actor}
\bibfield{author}{\bibinfo{person}{Junting Pan}, \bibinfo{person}{Siyu Chen}, \bibinfo{person}{Mike~Zheng Shou}, \bibinfo{person}{Yu Liu}, \bibinfo{person}{Jing Shao}, {and} \bibinfo{person}{Hongsheng Li}.} \bibinfo{year}{2021}\natexlab{}.
\newblock \showarticletitle{Actor-context-actor relation network for spatio-temporal action localization}. In \bibinfo{booktitle}{\emph{Proceedings of the IEEE/CVF Conference on Computer Vision and Pattern Recognition}}. \bibinfo{pages}{464--474}.
\newblock


\bibitem[Peng and Schmid(2016)]%
        {peng2016multi}
\bibfield{author}{\bibinfo{person}{Xiaojiang Peng} {and} \bibinfo{person}{Cordelia Schmid}.} \bibinfo{year}{2016}\natexlab{}.
\newblock \showarticletitle{Multi-region two-stream R-CNN for action detection}. In \bibinfo{booktitle}{\emph{Computer Vision--ECCV 2016: 14th European Conference, Amsterdam, The Netherlands, October 11--14, 2016, Proceedings, Part IV 14}}. Springer, \bibinfo{pages}{744--759}.
\newblock


\bibitem[Pramono et~al\mbox{.}(2019)]%
        {pramono2019hierarchical}
\bibfield{author}{\bibinfo{person}{Rizard Renanda~Adhi Pramono}, \bibinfo{person}{Yie-Tarng Chen}, {and} \bibinfo{person}{Wen-Hsien Fang}.} \bibinfo{year}{2019}\natexlab{}.
\newblock \showarticletitle{Hierarchical self-attention network for action localization in videos}. In \bibinfo{booktitle}{\emph{Proceedings of the IEEE/CVF international conference on computer vision}}. \bibinfo{pages}{61--70}.
\newblock


\bibitem[Qiu et~al\mbox{.}(2017)]%
        {qiu2017learning}
\bibfield{author}{\bibinfo{person}{Zhaofan Qiu}, \bibinfo{person}{Ting Yao}, {and} \bibinfo{person}{Tao Mei}.} \bibinfo{year}{2017}\natexlab{}.
\newblock \showarticletitle{Learning spatio-temporal representation with pseudo-3d residual networks}. In \bibinfo{booktitle}{\emph{proceedings of the IEEE International Conference on Computer Vision}}. \bibinfo{pages}{5533--5541}.
\newblock


\bibitem[Ren et~al\mbox{.}(2016)]%
        {ren2016faster}
\bibfield{author}{\bibinfo{person}{Shaoqing Ren}, \bibinfo{person}{Kaiming He}, \bibinfo{person}{Ross Girshick}, {and} \bibinfo{person}{Jian Sun}.} \bibinfo{year}{2016}\natexlab{}.
\newblock \showarticletitle{Faster R-CNN: Towards real-time object detection with region proposal networks}.
\newblock \bibinfo{journal}{\emph{IEEE transactions on pattern analysis and machine intelligence}} \bibinfo{volume}{39}, \bibinfo{number}{6} (\bibinfo{year}{2016}), \bibinfo{pages}{1137--1149}.
\newblock


\bibitem[Rong et~al\mbox{.}(2018)]%
        {rong2018finite}
\bibfield{author}{\bibinfo{person}{Nannan Rong}, \bibinfo{person}{Zhanshan Wang}, {and} \bibinfo{person}{Huaguang Zhang}.} \bibinfo{year}{2018}\natexlab{}.
\newblock \showarticletitle{Finite-time stabilization for discontinuous interconnected delayed systems via interval type-2 T--S fuzzy model approach}.
\newblock \bibinfo{journal}{\emph{IEEE Transactions on Fuzzy Systems}} \bibinfo{volume}{27}, \bibinfo{number}{2} (\bibinfo{year}{2018}), \bibinfo{pages}{249--261}.
\newblock


\bibitem[Rubio-Solis et~al\mbox{.}(2020)]%
        {rubio2020multilayer}
\bibfield{author}{\bibinfo{person}{Adrian Rubio-Solis}, \bibinfo{person}{George Panoutsos}, \bibinfo{person}{Carlos Beltran-Perez}, {and} \bibinfo{person}{Uriel Martinez-Hernandez}.} \bibinfo{year}{2020}\natexlab{}.
\newblock \showarticletitle{A multilayer interval type-2 fuzzy extreme learning machine for the recognition of walking activities and gait events using wearable sensors}.
\newblock \bibinfo{journal}{\emph{Neurocomputing}}  \bibinfo{volume}{389} (\bibinfo{year}{2020}), \bibinfo{pages}{42--55}.
\newblock


\bibitem[Simonyan and Zisserman(2014)]%
        {simonyan2014two}
\bibfield{author}{\bibinfo{person}{Karen Simonyan} {and} \bibinfo{person}{Andrew Zisserman}.} \bibinfo{year}{2014}\natexlab{}.
\newblock \showarticletitle{Two-stream convolutional networks for action recognition in videos}.
\newblock \bibinfo{journal}{\emph{Advances in neural information processing systems}}  \bibinfo{volume}{27} (\bibinfo{year}{2014}).
\newblock


\bibitem[Singh et~al\mbox{.}(2023)]%
        {singh2023eval}
\bibfield{author}{\bibinfo{person}{Ashish Singh}, \bibinfo{person}{Michael~J Jones}, {and} \bibinfo{person}{Erik~G Learned-Miller}.} \bibinfo{year}{2023}\natexlab{}.
\newblock \showarticletitle{Eval: Explainable video anomaly localization}. In \bibinfo{booktitle}{\emph{Proceedings of the IEEE/CVF Conference on Computer Vision and Pattern Recognition}}. \bibinfo{pages}{18717--18726}.
\newblock


\bibitem[Soomro et~al\mbox{.}(2012)]%
        {soomro2012ucf101}
\bibfield{author}{\bibinfo{person}{Khurram Soomro}, \bibinfo{person}{Amir~Roshan Zamir}, {and} \bibinfo{person}{Mubarak Shah}.} \bibinfo{year}{2012}\natexlab{}.
\newblock \showarticletitle{UCF101: A dataset of 101 human actions classes from videos in the wild}.
\newblock \bibinfo{journal}{\emph{arXiv preprint arXiv:1212.0402}} (\bibinfo{year}{2012}).
\newblock


\bibitem[Su et~al\mbox{.}(2019)]%
        {su2019improving}
\bibfield{author}{\bibinfo{person}{Rui Su}, \bibinfo{person}{Wanli Ouyang}, \bibinfo{person}{Luping Zhou}, {and} \bibinfo{person}{Dong Xu}.} \bibinfo{year}{2019}\natexlab{}.
\newblock \showarticletitle{Improving action localization by progressive cross-stream cooperation}. In \bibinfo{booktitle}{\emph{Proceedings of the IEEE/CVF Conference on Computer Vision and Pattern Recognition}}. \bibinfo{pages}{12016--12025}.
\newblock


\bibitem[Sui et~al\mbox{.}(2023)]%
        {sui2023simple}
\bibfield{author}{\bibinfo{person}{Lin Sui}, \bibinfo{person}{Chen-Lin Zhang}, \bibinfo{person}{Lixin Gu}, {and} \bibinfo{person}{Feng Han}.} \bibinfo{year}{2023}\natexlab{}.
\newblock \showarticletitle{A simple and efficient pipeline to build an end-to-end spatial-temporal action detector}. In \bibinfo{booktitle}{\emph{Proceedings of the IEEE/CVF Winter Conference on Applications of Computer Vision}}. \bibinfo{pages}{5999--6008}.
\newblock


\bibitem[Sun et~al\mbox{.}(2018)]%
        {sun2018actor}
\bibfield{author}{\bibinfo{person}{Chen Sun}, \bibinfo{person}{Abhinav Shrivastava}, \bibinfo{person}{Carl Vondrick}, \bibinfo{person}{Kevin Murphy}, \bibinfo{person}{Rahul Sukthankar}, {and} \bibinfo{person}{Cordelia Schmid}.} \bibinfo{year}{2018}\natexlab{}.
\newblock \showarticletitle{Actor-centric relation network}. In \bibinfo{booktitle}{\emph{Proceedings of the European Conference on Computer Vision (ECCV)}}. \bibinfo{pages}{318--334}.
\newblock


\bibitem[Tang et~al\mbox{.}(2020)]%
        {tang2020asynchronous}
\bibfield{author}{\bibinfo{person}{Jiajun Tang}, \bibinfo{person}{Jin Xia}, \bibinfo{person}{Xinzhi Mu}, \bibinfo{person}{Bo Pang}, {and} \bibinfo{person}{Cewu Lu}.} \bibinfo{year}{2020}\natexlab{}.
\newblock \showarticletitle{Asynchronous interaction aggregation for action detection}. In \bibinfo{booktitle}{\emph{Computer Vision--ECCV 2020: 16th European Conference, Glasgow, UK, August 23--28, 2020, Proceedings, Part XV 16}}. Springer, \bibinfo{pages}{71--87}.
\newblock


\bibitem[Wang et~al\mbox{.}(2016)]%
        {wang2016temporal}
\bibfield{author}{\bibinfo{person}{Limin Wang}, \bibinfo{person}{Yuanjun Xiong}, \bibinfo{person}{Zhe Wang}, \bibinfo{person}{Yu Qiao}, \bibinfo{person}{Dahua Lin}, \bibinfo{person}{Xiaoou Tang}, {and} \bibinfo{person}{Luc Van~Gool}.} \bibinfo{year}{2016}\natexlab{}.
\newblock \showarticletitle{Temporal segment networks: Towards good practices for deep action recognition}. In \bibinfo{booktitle}{\emph{European conference on computer vision}}. Springer, \bibinfo{pages}{20--36}.
\newblock


\bibitem[Wu and Krahenbuhl(2021)]%
        {wu2021towards}
\bibfield{author}{\bibinfo{person}{Chao-Yuan Wu} {and} \bibinfo{person}{Philipp Krahenbuhl}.} \bibinfo{year}{2021}\natexlab{}.
\newblock \showarticletitle{Towards long-form video understanding}. In \bibinfo{booktitle}{\emph{Proceedings of the IEEE/CVF Conference on Computer Vision and Pattern Recognition}}. \bibinfo{pages}{1884--1894}.
\newblock


\bibitem[Wu et~al\mbox{.}(2020)]%
        {wu2020context}
\bibfield{author}{\bibinfo{person}{Jianchao Wu}, \bibinfo{person}{Zhanghui Kuang}, \bibinfo{person}{Limin Wang}, \bibinfo{person}{Wayne Zhang}, {and} \bibinfo{person}{Gangshan Wu}.} \bibinfo{year}{2020}\natexlab{}.
\newblock \showarticletitle{Context-aware rcnn: A baseline for action detection in videos}. In \bibinfo{booktitle}{\emph{Computer Vision--ECCV 2020: 16th European Conference, Glasgow, UK, August 23--28, 2020, Proceedings, Part XXV 16}}. Springer, \bibinfo{pages}{440--456}.
\newblock


\bibitem[Yang et~al\mbox{.}(2021)]%
        {yang2021beyond}
\bibfield{author}{\bibinfo{person}{Xitong Yang}, \bibinfo{person}{Haoqi Fan}, \bibinfo{person}{Lorenzo Torresani}, \bibinfo{person}{Larry~S Davis}, {and} \bibinfo{person}{Heng Wang}.} \bibinfo{year}{2021}\natexlab{}.
\newblock \showarticletitle{Beyond short clips: End-to-end video-level learning with collaborative memories}. In \bibinfo{booktitle}{\emph{Proceedings of the IEEE/CVF Conference on Computer Vision and Pattern Recognition}}. \bibinfo{pages}{7567--7576}.
\newblock


\bibitem[Zhao et~al\mbox{.}(2022)]%
        {zhao2022tuber}
\bibfield{author}{\bibinfo{person}{Jiaojiao Zhao}, \bibinfo{person}{Yanyi Zhang}, \bibinfo{person}{Xinyu Li}, \bibinfo{person}{Hao Chen}, \bibinfo{person}{Bing Shuai}, \bibinfo{person}{Mingze Xu}, \bibinfo{person}{Chunhui Liu}, \bibinfo{person}{Kaustav Kundu}, \bibinfo{person}{Yuanjun Xiong}, \bibinfo{person}{Davide Modolo}, {et~al\mbox{.}}} \bibinfo{year}{2022}\natexlab{}.
\newblock \showarticletitle{Tuber: Tubelet transformer for video action detection}. In \bibinfo{booktitle}{\emph{Proceedings of the IEEE/CVF Conference on Computer Vision and Pattern Recognition}}. \bibinfo{pages}{13598--13607}.
\newblock


\bibitem[Zhao et~al\mbox{.}(2023)]%
        {zhao2023modulation}
\bibfield{author}{\bibinfo{person}{Weiji Zhao}, \bibinfo{person}{Kefeng Huang}, {and} \bibinfo{person}{Chongyang Zhang}.} \bibinfo{year}{2023}\natexlab{}.
\newblock \showarticletitle{Modulation-Based Center Alignment and Motion Mining for Spatial Temporal Action Detection}. In \bibinfo{booktitle}{\emph{ICASSP 2023-2023 IEEE International Conference on Acoustics, Speech and Signal Processing (ICASSP)}}. IEEE, \bibinfo{pages}{1--5}.
\newblock


\bibitem[Zhou et~al\mbox{.}(2018)]%
        {zhou2018temporal}
\bibfield{author}{\bibinfo{person}{Bolei Zhou}, \bibinfo{person}{Alex Andonian}, \bibinfo{person}{Aude Oliva}, {and} \bibinfo{person}{Antonio Torralba}.} \bibinfo{year}{2018}\natexlab{}.
\newblock \showarticletitle{Temporal relational reasoning in videos}. In \bibinfo{booktitle}{\emph{Proceedings of the European conference on computer vision (ECCV)}}. \bibinfo{pages}{803--818}.
\newblock


\end{thebibliography}

\end{document}